\documentclass{article}
\usepackage{xcolor, subcaption, cite, url, graphicx}
\usepackage{makecell}
\usepackage[textwidth=7.25in]{geometry}
\begin{document}
\newcommand{\yell}[1]{{\color{red}{\textbf{#1}}}}
\title{Overton: A Data System for Monitoring and Improving\\ Machine-Learned Products}

\author{
Christopher R\'e\\
Apple
\and
Feng Niu\\
Apple
\and
Pallavi Gudipati\\
Apple
\and
Charles Srisuwananukorn\\
Apple
}

\maketitle
\begin{abstract}\noindent
We describe a system called Overton, whose main design goal is to
support engineers in building, monitoring, and improving production
machine learning systems. Key challenges engineers face are monitoring
fine-grained quality, diagnosing errors in sophisticated applications,
and handling contradictory or incomplete supervision data. Overton
automates the life cycle of model construction, deployment, and
monitoring by providing a set of novel high-level, declarative
abstractions. Overton's vision is to shift developers to these
higher-level tasks instead of lower-level machine learning tasks. In
fact, using Overton, engineers can build deep-learning-based
applications without writing any code in frameworks like
TensorFlow. For over a year, Overton has been used in production to
support multiple applications in both near-real-time applications and
back-of-house processing. In that time, Overton-based applications
have answered billions of queries in multiple languages and processed
trillions of records reducing errors $1.7-2.9\times$ versus production
systems.
\end{abstract}

\section{Introduction}

In the life cycle of many production machine-learning applications,
maintaining and improving deployed models is the dominant factor in
their total cost and effectiveness--much greater than the cost of {\em
  de novo} model construction. Yet, there is little tooling for model
life-cycle support. For such applications, a key task for supporting
engineers is to improve and maintain the quality in the face of
changes to the input distribution and new production features. This
work describes a new style of data management system called Overton
that provides abstractions to support the model life cycle by helping
build models, manage supervision, and monitor application
quality.\footnote{The name Overton is a nod to the {\it Overton
    Window}, a concept from political theory that describes the set of
  acceptable ideas in public discourse. A corollary of this belief is
  that if one wishes to move the ``center'' of current discourse, one
  must advocate for a radical approaches outside the current
  window. Here, our radical approach was to focus only on programming
  by supervision and to prevent ML engineers from using ML toolkits
  like TensorFlow or manually selecting deep learning architectures.}

Overton is used in both near-real-time and backend production
applications. However, for concreteness, our running example is a
product that answers factoid queries, such as {\it ``how tall is the
  president of the united states?''} In our experience, the engineers
who maintain such machine learning products face several challenges on
which they spend the bulk of their time.

\begin{itemize}
\item \textbf{Fine-grained Quality Monitoring} While overall
  improvements to quality scores are important, often the week-to-week
  battle is improving fine-grained quality for important subsets of
  the input data. An individual subset may be rare but are nonetheless
  important, e.g., 0.1\% of queries may correspond to a product
  feature that appears in an advertisement and so has an outsized
  importance. Traditional machine learning approaches effectively
  optimize for aggregate quality. As hundreds of such subsets are
  common in production applications, this presents data management and
  modeling challenges.  An ideal system would monitor these subsets
  and provide tools to improve these subsets while maintaining overall
  quality.
  
\item \textbf{Support for Multi-component Pipelines} Even simple
  machine learning products comprise myriad individual
  tasks. Answering even a simple factoid query, such as {\it ``how
    tall is the president of the united states?''}  requires tackling
  many tasks including (1) find the named entities (`united states',
  and `president'), (2) find the database ids for named entities, (3)
  find the intent of the question, e.g., the height of the topic
  entity, (4) determine the topic entity, e.g., neither president nor
  united states, but the person Donald J. Trump, who is not explicitly
  mentioned, and (5) decide the appropriate UI to render it on a
  particular device. Any of these tasks can go wrong. Traditionally,
  systems are constructed as pipelines, and so determining which task
  is the culprit is challenging.
  
\item \textbf{Updating Supervision} When new features are created or
  quality bugs are identified, engineers provide additional
  supervision. Traditionally, supervision is provided by annotators
  (of varying skill levels), but increasingly {\it programmatic
    supervision} is the dominant form of
  supervision~\cite{DBLP:conf/sigmod/BachRLLSXSRHAKR19,
    DBLP:conf/nips/RatnerSWSR16}, which includes labeling, data
  augmentation, and creating synthetic data. For both privacy and cost
  reasons, many applications are constructed using programmatic
  supervision as a primary source. An ideal system can accept
  supervision at multiple granularities and resolve conflicting
  supervision for those tasks.
\end{itemize}

There are other desiderata for such a system, but the commodity
machine learning stack has evolved to support them: building
deployment models, hyperparameter tuning, and simple model search are
now well supported by commodity packages including TensorFlow,
containers, and (private or public) cloud infrastructure.\footnote{Overton
builds on systems like
Turi~\cite{DBLP:conf/sigmod/AgrawalABBGGLMP19}.} By combining these new
systems, Overton is able to automate many of the traditional modeling
choices, including deep learning architecture, its hyperparameters,
and even which embeddings are used.

Overton provides the engineer with abstractions that allow them to
build, maintain, and monitor their application by manipulating data
files--not custom code. Inspired by relational systems, supervision
(data) is managed separately from the model (schema). Akin to
traditional {\em logical independence}, Overton's schema provides {\em
  model independence}: serving code does not change even when inputs,
parameters, or resources of the model change. The schema changes very
infrequently--many production services have not updated their schema
in over a year.

Overton takes as input a {\em schema} whose design goal is to support
rich applications from modeling to automatic deployment. In more
detail, the {\em schema} has two elements: (1) {\em data
  payloads} similar to a relational schema, which describe the input
data, and (2) {\em model tasks}, which describe the tasks that need to
be accomplished. The schema defines the input, output, and
coarse-grained data flow of a deep learning model. Informally, {\it
  the schema defines what the model computes but not how the model
  computes it}: Overton does not prescribe architectural details of
the underlying model (e.g., Overton is free to embed sentences using
an LSTM or a Transformer) or hyperparameters, like hidden state size.
Additionally, sources of supervision are described as data--not in the
schema--so they are free to rapidly evolve.

\begin{figure}
  \centering
  \includegraphics[width=0.475\textwidth]{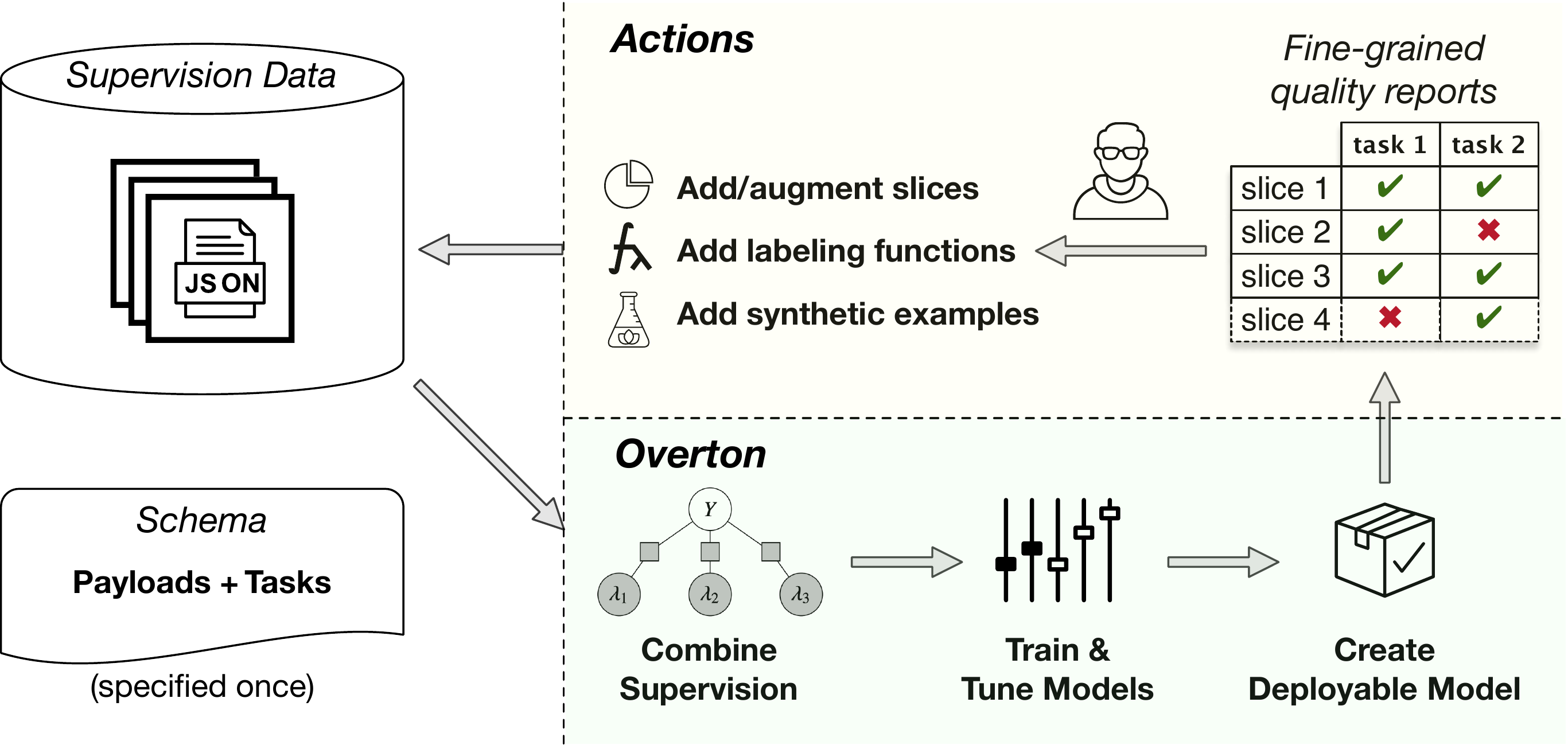}
  \caption{Schema and supervision data are input to Overton, which
    outputs a deployable model. Engineers monitor and improve the
    model via supervision data.}
  \label{fig:workflow}
\end{figure}

As shown in Figure~\ref{fig:workflow}, given a schema and a data file,
Overton is responsible to instantiate and train a model, combine
supervision, select the model's hyperparameters, and produce a
production-ready binary. Overton compiles the schema into a
(parameterized) TensorFlow or PyTorch program, and performs an
architecture and hyperparameter search. A benefit of this compilation
approach is that Overton can use standard toolkits to monitor training
(TensorBoard equivalents) and to meet service-level agreements
(Profilers). The models and metadata are written to an S3-like data
store that is accessible from the production infrastructure. This has
enabled model retraining and deployment to be nearly automatic,
allowing teams to ship products more quickly.

In retrospect, the following three choices of Overton were the most
important in meeting the above challenges.

\textbf{(1) Code-free Deep Learning} In Overton-based systems,
engineers focus exclusively on fine-grained monitoring of their
application quality and improving supervision--not tweaking deep
learning models. An Overton engineer does not write any deep learning
code in frameworks like TensorFlow. To support application quality
improvement, we use a technique, called {\em model
  slicing}~\cite{slicing}. The main idea is to allow the developer to
identify fine-grained subsets of the input that are important to the
product, e.g., queries about nutrition or queries that require
sophisticated disambiguation. The system uses developer-defined slices
as a guide to increase representation capacity. Using this recently
developed technique led to state-of-the-art results on natural
language benchmarks including GLUE and
SuperGLUE~\cite{DBLP:conf/emnlp/WangSMHLB18}.\footnote{A blog post
  introduction to slicing is here \url{https://snorkel.org/superglue.html}.}

  \textbf{(2) Multitask Learning} Overton was built to natively
  support multitask
  learning~\cite{Caruana93multitasklearning,DBLP:journals/corr/Ruder17a,sogaard2016deep}
  so that all model tasks are concurrently predicted. A key benefit is
  that Overton can accept supervision at whatever granularity (for
  whatever task) is available. Overton models often perform ancillary
  tasks like part-of-speech tagging or typing. Intuitively, if a
  representation has captured the semantics of a query, then it should
  reliably perform these ancillary tasks. Typically, ancillary tasks
  are also chosen either to be inexpensive to supervise. Ancillary
  task also allow developers to gain confidence in the model's
  predictions and have proved to be helpful for aids for debugging errors.

\textbf{(3) Weak Supervision} Applications have access to supervision
of varying quality and combining this contradictory and incomplete
supervision is a major challenge. Overton uses techniques from
Snorkel~\cite{DBLP:conf/nips/RatnerSWSR16} and Google's Snorkel
DryBell~\cite{DBLP:conf/sigmod/BachRLLSXSRHAKR19}, which have studied
how to combine supervision in theory and in software. Here, we
describe two novel observations from building production applications:
(1) we describe the shift to applications which are constructed almost
{\it entirely} with weakly supervised data due to cost, privacy, and
cold-start issues, and (2) we observe that weak supervision may
obviate the need for popular methods like transfer learning from
massive pretrained models, e.g.,
BERT~\cite{DBLP:conf/naacl/DevlinCLT19}--on some production workloads,
which suggests that a deeper trade-off study may be illuminating.

In summary, Overton represents a first-of-its kind machine-learning
lifecycle management system that has a focus on monitoring and
improving application quality. A key idea is to separate the model and
data, which is enabled by a code-free approach to deep
learning. Overton repurposes ideas from the database community and the
machine learning community to help engineers in supporting the
lifecycle of machine learning toolkits. This design is informed and
refined from use in production systems for over a year in multiple
machine-learned products.

\section{An Overview of Overton}

\begin{figure*}[ht]
  \centering
  \begin{subfigure}{0.45\textwidth}\centering
  \includegraphics[width=\textwidth]{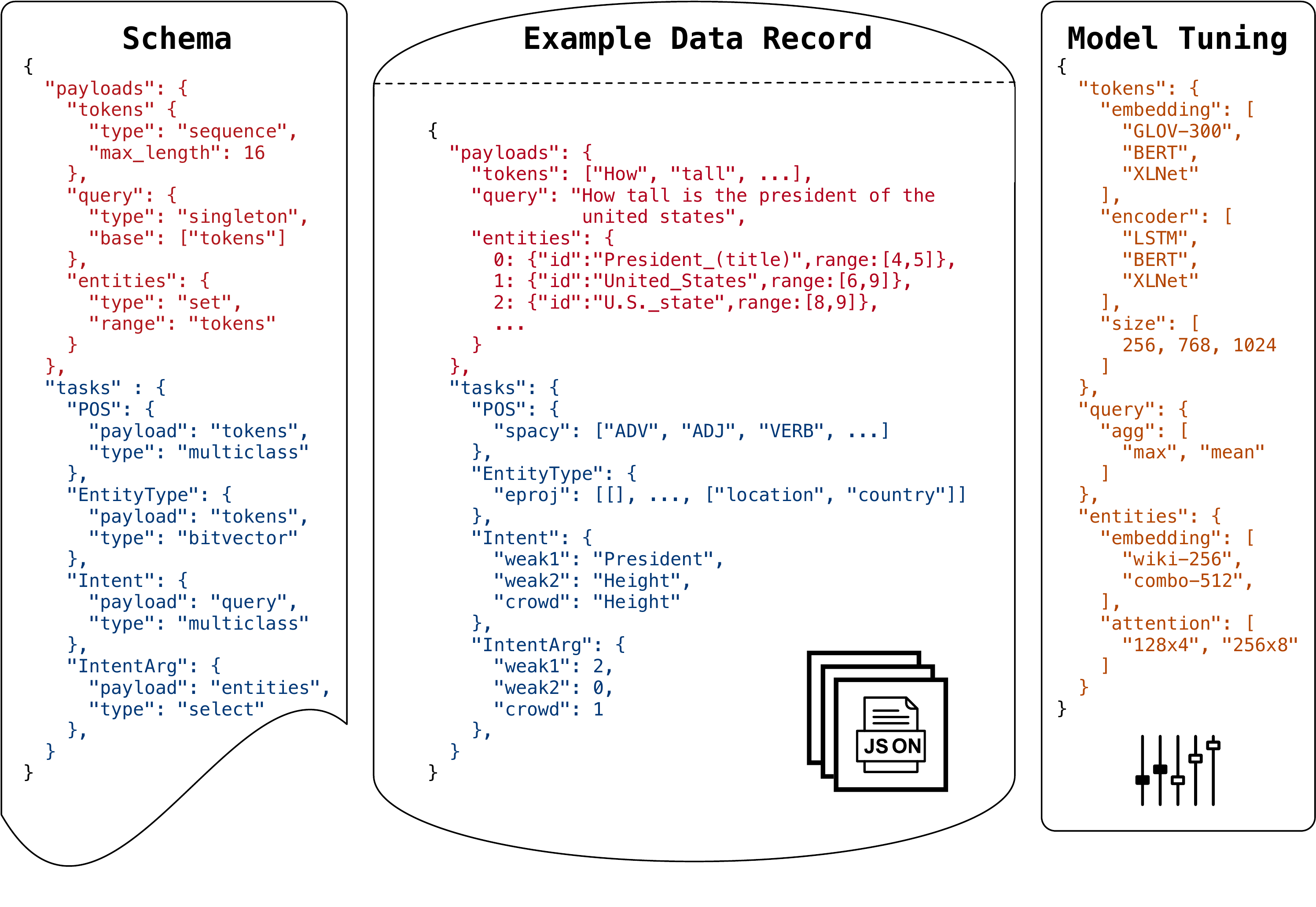}
  \caption{An example schema, data file, and tuning specification for
    our running example. Colors group logical sections.  }
  \label{fig:schema}
  \end{subfigure}\hfill%
  \begin{subfigure}{0.45\textwidth}\centering
  \includegraphics[width=\textwidth]{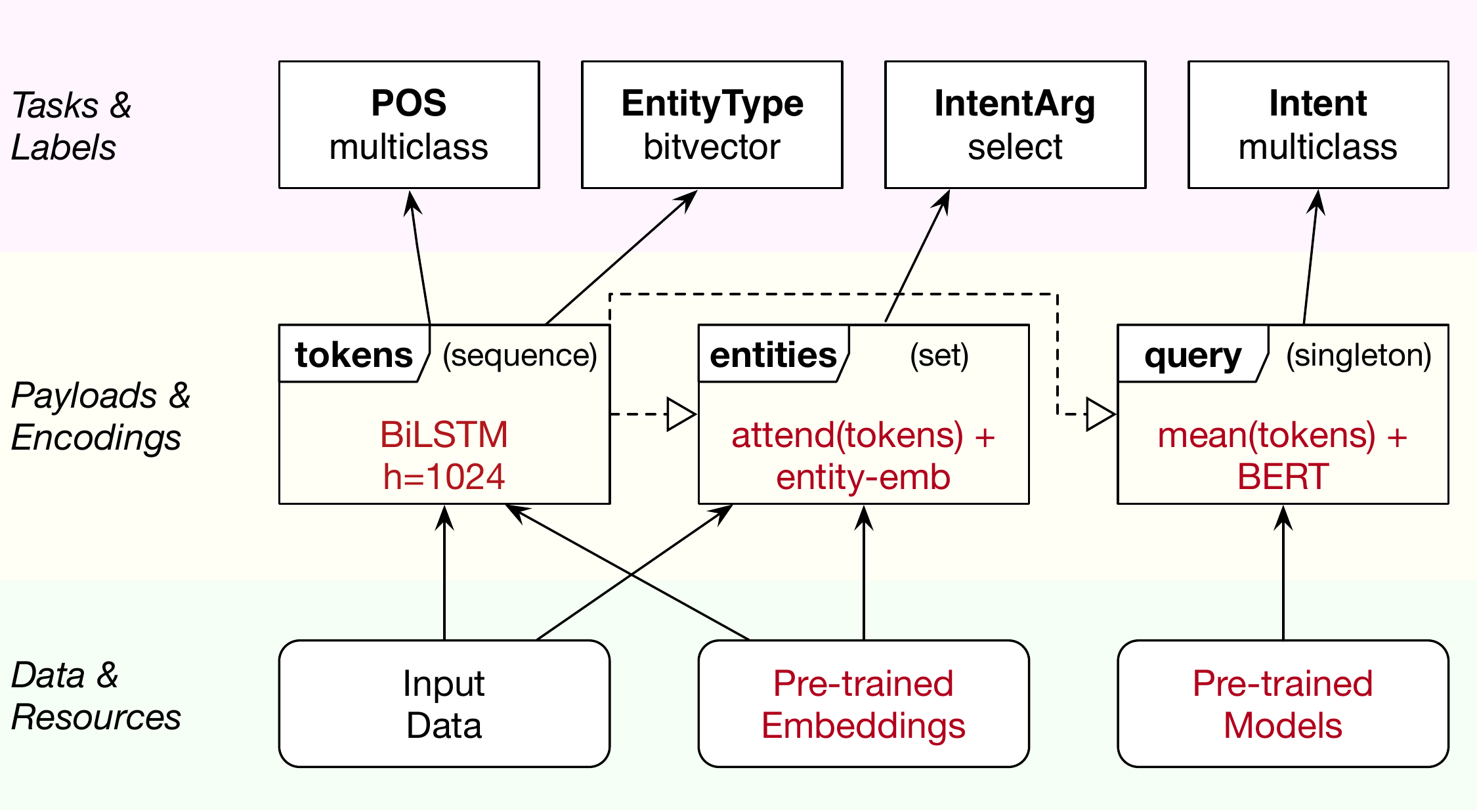}
  \caption{A deep architecture that might be selected by Overton. The
    red components are selected by Overton via model search; the black boxes and
    connections are defined by the schema.}
  \label{fig:network}
  \end{subfigure}
  \caption{The inputs to Overton and a schematic view of a compiled
    model.  }
\end{figure*}

To describe the components of Overton, we continue our running example
of a factoid answering product. Given the textual version of a
query, e.g., {\it ``how tall is the president of the united
  states''}, the goal of the system is to appropriately render the
answer to the query. The main job of an engineer is to measure and
improve the quality of the system across many queries, and a key
capability Overton needs to support is to measure the quality in
several fine-grained ways. This quality is measured within Overton by
evaluation on curated test sets, which are fastidiously maintained and
improved by annotators and engineers. An engineer may be responsible
for improving performance on a specific subset of the data, which they
would like to monitor and improve.

There are two inputs to Overton (Figure~\ref{fig:schema}): The schema
(Section~\ref{subsec:schema}), which specifies the tasks, and a data
file, which is the primary way an engineer refines quality
(Section~\ref{subsec:slices}). Overton then compiles these inputs into
a multitask deep model (Figure~\ref{fig:network}). We describe an
engineer's interaction with Overton (Section~\ref{subsec:eng}) and
discuss design decisions (Section~\ref{subsec:design}).

\subsection{Overton's Schema}
\label{subsec:schema}

An Overton schema has two components: the {\em tasks}, which capture
the tasks the model needs to accomplish, and {\em payloads}, which
represent sources of data, such as tokens or entity embeddings. Every
example in the data file conforms to this schema. Overton uses a
schema both as a guide to compile a TensorFlow model and to describe
its output for downstream use.\footnote{Overton also supports PyTorch
  and CoreML backends. For brevity, we describe only TensorFlow.}
Although Overton supports more types of tasks, we focus on
classification tasks for simplicity. An example schema and its
corresponding data file are shown in Figure~\ref{fig:schema}. The
schema file also provides schema information in a traditional database
sense: it is used to define a memory-mapped row-store for
example.\footnote{Since all elements of an example are needed
  together, a row store has obvious IO benefits over column-store-like
  solutions.}

A key design decision is that the schema does not contain information
about hyperparameters like hidden state sizes. This enables {\em model
  independence}: the same schema is used in many downstream
applications and even across different languages. Indeed, the same
schema is shared in multiple locales and applications, only the
supervision differs.

\paragraph*{Payloads}
Conceptually, Overton embeds raw data into a payload, which is then
used as input to a task or to another payload. Overton supports
payloads that are singletons (e.g., a query), sequences (e.g. a query
tokenized into words or characters), and sets (e.g., a set of
candidate entities). Overton's responsibility is to embed these
payloads into tensors of the correct size, e.g., a query is embedded
to some dimension $d$, while a sentence may be embedded into an array
of size $m \times d$ for some length $m$. The mapping from inputs can
be learned from scratch, pretrained, or fine-tuned; this allows
Overton to incorporate information from a variety of different sources
in a uniform way.

Payloads may refer directly to a data field in a record for input,
e.g., a field `tokens' contains a tokenized version of the
query. Payloads may also refer to the contents of another payload. For
example, a query payload may aggregate the representation of all
tokens in the query. A second example is that an entity payload may
refer to its corresponding span of text, e.g., the ``united states of
america'' entity points to the span ``united states'' in the
query. Payloads may aggregate several sources of information by
referring to a combination of source data and other payloads. The
payloads simply indicate dataflow, Overton learns the semantics of
these references.\footnote{ By default, combination is done with
  multi-headed attention. The method of aggregation is not specified
  and Overton is free to change it.}

\paragraph*{Tasks}
Continuing our running example in Figure~\ref{fig:network}, we see
four tasks that refer to three different payloads. For each payload
type, Overton defines a multiclass and a bitvector classification
task. In our example, we have a multiclass model for the
\textsc{intent} task: it assigns one label for each query payload,
e.g., the query is about ``height''. In contrast, in the
\textsc{EntityType} task, fine-grained types for each token are not
modeled as exclusive, e.g., location and country are not
exclusive. Thus, the \textsc{EntityType} task takes the token payloads
as input, and emits a bitvector for each token as output. Overton also
supports a task of selecting one out of a set, e.g.,
\textsc{IntentArg} selects one of the candidate entities. This
information allows Overton to compile the inference code and the loss
functions for each task and to build a {\em serving signature}, which
contains detailed information of the types and can be consumed by
model serving infrastructure. At the level of TensorFlow, Overton
takes the embedding of the payload as input, and builds an output
prediction and loss function of the appropriate type.

The schema is changed infrequently, and many engineers who use Overton
simply select an existing schema. Applications are customized by
providing supervision in a data file that conforms to the schema,
described next.

\subsection{Weak Supervision and Slices}
\label{subsec:slices}
The second main input to Overton is the data file. It is specified as
(conceptually) a single file: the file is meant to be engineer
readable and queryable (say using jq), and each line is a single JSON
record. For readability, we have pretty-printed a data record in
Figure~\ref{fig:schema}. Each payload is described in the file (but
may be null).

The supervision is described under each task, e.g., there are three
(conflicting) sources for the \textsc{Intent} task.  A task requires
labels at the appropriate granularity (singleton, sequence, or set)
and type (multiclass or bitvector). The labels are tagged by the
source that produced them: these labels may be incomplete and even
contradictory. Overton models the sources of these labels, which may
come human annotators, or from engineer-defined heuristics such as
data augmentation or heuristic labelers. Overton learns the accuracy of these
sources using ideas from the Snorkel
project~\cite{DBLP:conf/nips/RatnerSWSR16}. In particular, it
estimates the accuracy of these sources and then uses these accuracies
to compute a probability that each training point is
correct~\cite{DBLP:conf/icml/VarmaSHRR19}. Overton incorporates this
information into the loss function for a task; this also allows
Overton to automatically handle common issues like rebalancing
classes.

\paragraph*{Monitoring} For monitoring, Overton allows engineers to provide
user-defined {\em tags} that are associated with individual data
points. The system additionally defines default tags including {\em
  train}, {\em test}, {\em dev} to define the portion of the data that
should be used for training, testing, and development. Engineers are
free to define their own subsets of data via tags, e.g,. the date
supervision was introduced, or by what method. Overton allows report
per-tag monitoring, such as the accuracy, precision and recall, or
confusion matrices, as appropriate. These tags are stored in a format
that is compatible with Pandas. As a result, engineers can load these
tags and the underlying examples into other downstream analysis tools
for further analytics.

\paragraph{Slicing} In addition to tags, Overton defines a mechanism
called {\em slicing}, that allows monitoring but also adds
representational capacity to the model.  An engineer defines a slice
by tagging a subset of the data and indicating that this tag is also a
slice. Engineers typically define slices that consist of a subset that
is particular relevant for their job. For example, they may define a
slice because it contains related content, e.g., {\it
  ``nutrition-related queries''} or because the subset has an
interesting product feature, e.g., {\it ``queries with complex
  disambiguation''}. The engineer interacts with Overton by
identifying these slices, and providing supervision for examples in
those slices.\footnote{Downstream systems have been developed to
  manage the slicing and programmatic supervision from the UI
  perspective that are managed by independent teams.}  Overton reports
the accuracy conditioned on an example being in the slice. The main
job of the engineer is to diagnose what kind of supervision would
improve a slice, and refine the labels in that slice by correcting
labels or adding in new labels.

A slice also indicates to Overton that it should increase its
representation capacity (slightly) to learn a {\it ``per slice''}
representation for a task.\footnote{We only describe its impact on
  systems architecture, the machine learning details are described in
  Chen et al.~\cite{slicing}.} In this sense, a slice is akin to
defining a ``micro-task'' that performs the task just on the subset
defined by the slice. Intuitively, this slice should be able to better
predict as the data in a slice typically has less variability than the
overall data. At inference time, Overton makes only one prediction per
task, and so the first challenge is that Overton needs to combine
these overlapping slice-specific predictions into a single prediction.
A second challenge is that slices heuristically (and so imperfectly)
define subsets of data. To improve the coverage of these slices,
Overton learns a representation of when one is {\it ``in the slice''}
which allows a slice to generalize to new examples.  Per-slice
performance is often valuable to an engineer, even if it does not
improve the overall quality, since their job is to improve and monitor
a particular slice. A production system improved its performance on a
slice of complex but rare disambiguations by over $50$ points of F1
using the same training data.

\subsection{A Day in the Life of an Overton Engineer}
\label{subsec:eng}

To help the reader understand the process of an engineer, we describe
two common use cases: improving an existing feature, and the
cold-start case. Overton's key ideas are changing where developers
spend their time in this process.

\paragraph*{Improving an Existing Feature}
A first common use case is that an engineer wants to improve the
performance of an existing feature in their application. The developer
iteratively examines logs of the existing application. To support this
use case, there are downstream tools that allow one to quickly define
and iterate on subsets of data. Engineers may identify areas of the
data that require more supervision from annotators, conflicting
information in the existing training set, or the need to create new
examples through weak supervision or data augmentation. Over time,
systems have grown on top of Overton that support each of these
operations with a more convenient UI. An engineer using Overton may
simply work entirely in these UIs.

\paragraph*{Cold-start Use Case}
A second common use case is the cold-start use case. In this case, a
developer wants to launch a new product feature. Here, there is no
existing data, and they may need to develop synthetic data. In both
cases, the identification and creation of the subset is done by tools
outside of Overton. These subsets become the aforementioned slices,
and the different mechanisms are identified as different
sources. Overton supports this process by allowing engineers to tag
the lineage of these newly created queries, measure their quality in a
fine-grained way, and merge data sources of different quality.\\

In previous iterations, engineers would modify loss functions by hand
or create new separate models for each case.  Overton engineers spend
no time on these activities. 

\subsection{Major Design Decisions and Lessons}
\label{subsec:design}
We briefly cover some of the design decisions in Overton.

\paragraph*{Design for Weakly Supervised Code}
As described, weakly supervised machine learning is often the dominant
source of supervision in many machine learning products. Overton uses
ideas from Snorkel~\cite{DBLP:conf/nips/RatnerSWSR16} and Google's
Snorkel Drybell~\cite{DBLP:conf/sigmod/BachRLLSXSRHAKR19} to model the
quality of the supervision. The design is simple: lineage is tracked
for each source of information. There are production systems with {\em
  no} traditional supervised training data (but they do have such data
for validation). This is important in privacy-conscious applications.

\paragraph*{Modeling to Deployment}
In many production teams, a deployment team is distinct from the
modeling team, and the deployment team tunes models for
production. However, we noticed quality regressions as deployment
teams have an incomplete view of the potential modeling
tradeoffs. Thus, Overton was built to construct a deployable
production model. The runtime performance of the model is potentially
suboptimal, but it is well within production SLAs. By encompassing
more of the process, Overton has allowed faster model turn-around
times.

\paragraph*{Use Standard Tools for the ML Workflow}
Overton compiles the schema into (many versions of) TensorFlow,
CoreML, or PyTorch. Whenever possible, Overton uses a standard
toolchain. Using standard tools, Overton supports distributed
training, hyperparameter tuning, and building servable models. One
unanticipated benefit of having both backends was that different
resources are often available more conveniently on different
platforms. For example, to experiment with pretrained models, the
Huggingface repository~\cite{huggingface} allows quick
experimentation--but only in PyTorch. The TensorFlow production tools
are unmatched. The PyTorch execution mode also allows REPL and
in-Jupyter-notebook debugging, which engineers use to repurpose
elements, e.g., query similarity features. Even if a team uses a
single runtime, different runtime services will inevitably use
different versions of that runtime, and Overton insulates the modeling
teams from the underlying changes in production serving
infrastructure.

\paragraph*{Model Independence and Zero-code Deep Learning}
A major design choice at the outset of the project was that domain
engineers should not be forced to write traditional deep learning
modeling code. Two years ago, this was a contentious decision as the
zeitgeist was that new models were frequently published, and this
choice would hamstring the developers. However, as the pace of new
model building blocks has slowed, domain engineers no longer feel the
need to fine-tune individual components at the level of
TensorFlow. Ludwig\footnote{\url{https://uber.github.io/ludwig/}} has
taken this approach and garnered adoption. Although developed
separately, Overton's schema looks very similar to Ludwig's programs
and from conversations with the developers, shared similar
motivations. Ludwig, however, focused on the one-off model building
process not the management of the model lifecycle. Overton itself only
supports text processing, but we are prototyping image, video, and
multimodal applications.

\paragraph*{Engineers are Comfortable with Automatic Hyperparameter Tuning}
Hyperparameter tuning is conceptually important as it allows Overton
to avoid specifying parameters in the schema for the model
builder. Engineers are comfortable with automatic tuning, and first
versions of all Overton systems are tuned using standard
approaches. Of course, engineers may override the search: Overton is
used to produce servable models, and so due to SLAs, production models
often pin certain key parameters to avoid tail performance
regressions.

\paragraph*{Make it easy to manage ancillary data products}
Overton is also used to produce back-end data products (e.g., updated
word or multitask embeddings) and multiple versions of the same
model. Inspired by HuggingFace~\cite{huggingface}, Overton tries to
make it easy to drop in new pretrained embeddings as they arrive: they
are simply loaded as payloads. Teams use multiple models to train a
``large'' and a ``small'' model on the same data. The large model is
often used to populate caches and do error analysis, while the small
model must meet SLA requirements. Overton makes it easy to keep these
two models synchronized. Additionally, some data products can be
expensive to produce (on the order of ten days), which means they are
refreshed less frequently than the overall product. Overton does not
have support for model versioning, which is likely a design oversight.

\section{Evaluation}

We elaborate on three items: (1) we describe how Overton improves
production systems; (2) we report on the use of weak supervision in
these systems; and (3) we discuss our experience with
pretraining.\footnote{Due to sensitivity around production systems, we
  report relative quality numbers and obfuscate some tasks.}

\paragraph*{Overton Usage}
Overton has powered industry-grade systems for more than a year. Figure~\ref{table:prod} shows the end-to-end reduction in error of these systems:  a high-resource system with tens of engineers, a large budget, and large existing training sets, and three other products with smaller teams.  Overton enables a small team to perform the same duties that would traditionally be done by several, larger teams.  Here, multitask learning is critical:  the combined system reduces error and improves product turn-around times. Systems that Overton models replace are typically deep models and heuristics that are challenging to maintain, in our estimation because there is no model independence.

\begin{figure}
  \centering
  \begin{tabular}{cc||c}
      \thead{Resourcing} & \thead{Error Reduction} & \thead{Amount of Weak Supervision}\\
      \hline
      High & 65\% $(2.9\times)$ & 80\% \\
      Medium & 82\% $(5.6\times)$ & 96\% \\
      Medium & 72\% $(3.6\times)$ & 98\% \\
      Low & 40\% $(1.7\times)$ & 99\% \\
  \end{tabular}
  \caption{For products at various resource levels, percentage (and factor) fewer errors of Overton system makes compared to previous system, and the percentage of weak supervision of all supervision.}
  \label{table:prod}
\end{figure}

\paragraph*{Usage of Weak Supervision}
Weak supervision is the dominant form of supervision in all
applications. Even annotator labels (when used) are filtered and
altered by privacy and programmatic quality control steps. Note that
{\em validation} is still done manually, but this requires orders of
magnitude less data than training.

\begin{figure}[h]
    \centering
    \includegraphics[width=0.48\textwidth]{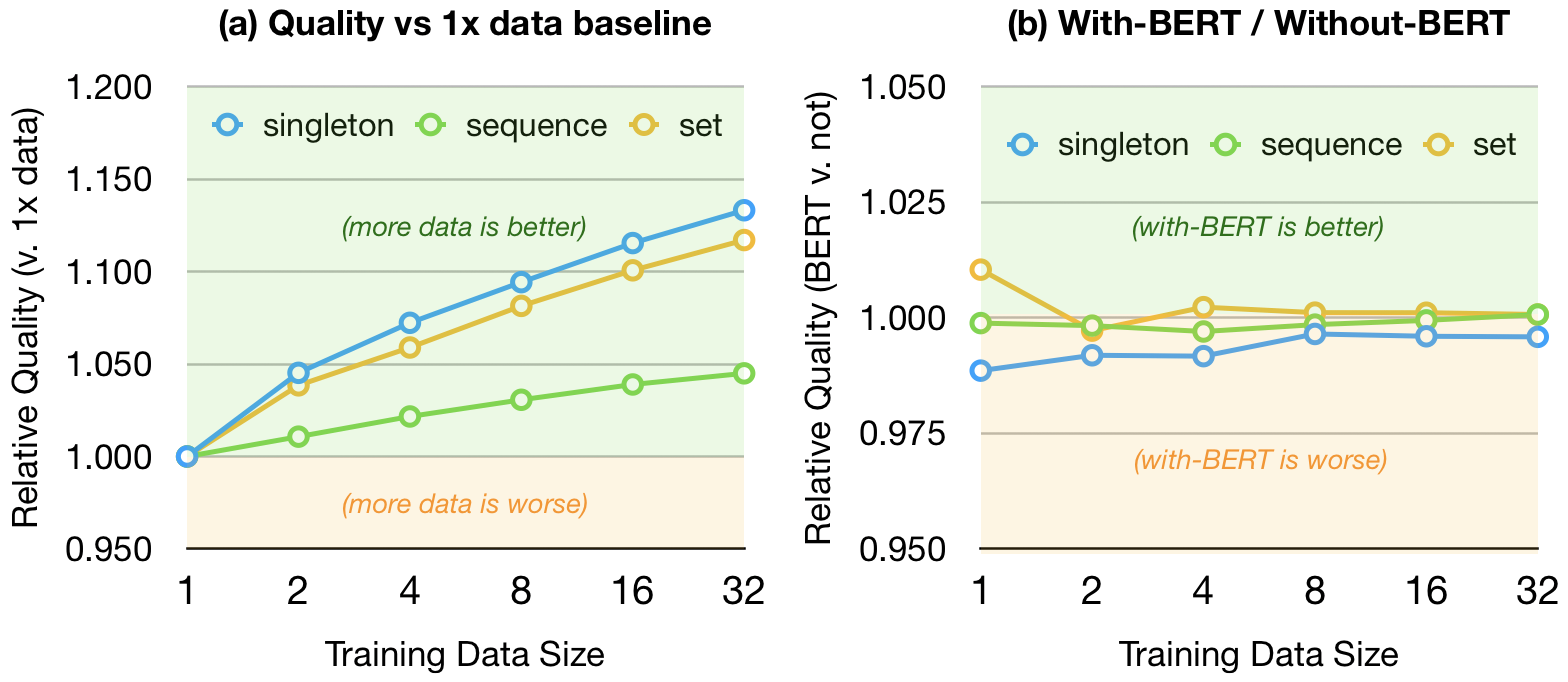}
    \caption{Relative quality changes as data set size scales.}
    \label{fig:relative:quality}
\end{figure}

Figure~\ref{fig:relative:quality}a shows the impact of weak
supervision on quality versus weak supervision scale. We downsample
the training data and measure the test quality (F1 and accuracy) on 3
representative tasks: singleton, sequence, and set.\footnote{We
  obfuscate tasks using their underlying payload type.}  For each
task, we use the 1x data's model as the baseline and plot the relative
quality as a percentage of the baseline; e.g., if the baseline F1 is
0.8 and the subject F1 is 0.9, the relative quality is
$0.9/0.8=1.125$. In Figure~\ref{fig:relative:quality}a, we see that
increasing the amount of supervision consistently results in improved
quality across all tasks. Going from 30K examples or so (1x) to 1M
examples (32x) leads to a 12\%+ bump in two tasks and a 5\% bump in
one task.

\paragraph*{Pre-trained Models and Weak Supervision}
A major trend in the NLP community is to pre-train a large and complex
language model using raw text and then fine-tune it for specific
tasks~\cite{DBLP:conf/naacl/DevlinCLT19}. One can easily integrate
such pre-trained models in Overton, and we were excited by our early
results. Of course, at some point, training data related to the task
is more important than massive pretraining. We wondered how weak
supervision and pretrained models would interact. Practically, these
pretrained models like BERT take large amounts of memory and are much
slower than standard word embeddings. Nevertheless, motivated by such
models' stellar performance on several recent NLP benchmarks such as
GLUE~\cite{DBLP:conf/emnlp/WangSMHLB18}, we evaluate their impact on
production tasks that are weakly supervised. For each of the
aforementioned training set sizes, we train two models:
\textbf{without-BERT}: production model with standard word embeddings
but without BERT, and \textbf{with-BERT}: production model with fine
tuning on the ``BERT-Large, Uncased'' pretrained
model~\cite{DBLP:conf/naacl/DevlinCLT19}.

For each training set, we calculate the relative test quality change
(percentage change in F1 or accuracy) of \textbf{with-BERT} over
\textbf{without-BERT}.  In Figure~\ref{fig:relative:quality}b, almost
all percentage changes are within a narrow 2\% band of no-change
(i.e., 100\%). This suggests that sometimes pre-trained language
models have a limited impact on downstream tasks--when weak
supervision is used.  Pretrained models do have higher quality at
smaller training dataset sizes--the Set task here shows an improvement
at small scale, but this advantage vanishes at larger (weak) training
set sizes in these workloads. This highlights a potentially
interesting set of tradeoffs among weak supervision, pretraining, and
the complexity of models.

\section{Related Work}

Overton builds on work in model life-cycle management, weak
supervision, software for ML, and zero-code deep learning.

\paragraph*{Model Management}
A host of recent data systems help manage the model process, including
MLFlow\footnote{\url{https://mlflow.org}}, which helps with the model lifecycle and
reporting~\cite{DBLP:journals/debu/ZahariaCD0HKMNO18},
ModelDB~\cite{DBLP:journals/debu/VartakM18}, and more. Please see
excellent tutorials such as Kumar et
al.~\cite{DBLP:conf/sigmod/Kumar0017}. However, these
systems are complementary and do not focus on Overton's three design
points: fine-grained monitoring, diagnosing the workflow of updating
supervision, and the production programming lifecycle. This paper
reports on some key lessons learned from productionizing related ideas.

\paragraph*{Weak Supervision} A myriad of weak supervision techniques have been used over the last
few decades of machine learning, notably external knowledge
bases~\cite{mintz2009distant,zhang:cacm17,craven:ismb99,takamatsu:acl12},
heuristic patterns~\cite{gupta2014improved,ratner2018snorkel}, feature
annotations~\cite{mann2010generalized,zaidan:emnlp08}, and noisy crowd
labels~\cite{karger2011iterative,dawid1979maximum}. Data augmentation
is another major source of training data. One promising approach is to
learn augmentation policies, first described in Ratner et
al.~\cite{DBLP:conf/nips/RatnerEHDR17}, which can further automate
this process. Google's
AutoAugment~\cite{DBLP:journals/corr/abs-1805-09501} used learned
augmentation policies to set new state-of-the-art performance results
in a variety of domains, which has been a tremendously exciting
direction. The goal of systems like Snorkel is to unify and extend
these techniques to create and manipulate training
data.\footnote{\url{http://snorkel.org}} These have recently garnered
usage at major companies, notably Snorkel DryBell at
Google~\cite{DBLP:conf/sigmod/BachRLLSXSRHAKR19}. Overton is inspired
by this work and takes the next natural step toward supervision
management.

\paragraph*{Software Productivity for ML Software}
The last few years have seen an unbelievable amount of change in the
machine learning software landscape. TensorFlow, PyTorch, CoreML and
MXNet have changed the way people write machine learning code to build
models. Increasingly, there is a trend toward higher level
interfaces. The pioneering work on higher level domain specific
languages like Keras\footnote{\url{http://keras.io}} began in this
direction. Popular libraries like Fast.ai, which created a set of
libraries and training materials, have dramatically improved engineer
productivity.\footnote{\url{http://fast.ai}} These resources have made
it easier to build models but equally important to train model
developers. Enabled in part by this trend, Overton takes a different
stance: {\em model development is in some cases not the key to product
  success}. Given a fixed budget of time to run a long-lived ML model,
Overton is based on the idea that success or failure depends on
engineers being able to iterate quickly and maintain the
supervision--not change the model. Paraphrasing the classical
relational database management mantra, Overton focuses on what the
user wants--not how to get it.

\paragraph*{Zero-code Deep Learning} The ideas above led naturally to what we now recognize as {\it zero-code
  deep learning}, a term we borrow from Ludwig. It is directly related
to previous work on multitask learning as a key building block of
software development~\cite{DBLP:conf/cidr/RatnerHR19} and inspired by
Software 2.0 ideas articulated by
Karpathy.\footnote{\url{https://medium.com/@karpathy/software-2-0-a64152b37c35}.}
The world of software engineering for machine learning is fascinating
and nascent. In this spirit, Uber's Ludwig shares a great deal with
Overton's design. Ludwig is very sophisticated and has supported
complex tasks on vision and others. These methods were controversial
two years ago, but seem to be gaining acceptance among production
engineers. For us, these ideas began as an extension of joint
inference and learning in
DeepDive~\cite{DBLP:journals/pvldb/ShinWWSZR15}.

\paragraph*{Network Architecture Search}
Zero-code deep learning in Overton is enabled by some amount of
architecture search. It should be noted that Ludwig made a different
choice: no search is required, and so zero-code deep learning does not
depend on search. The area of Neural Architecture Search
(NAS)~\cite{DBLP:journals/jmlr/ElskenMH19} is booming: the goal of
this area is to perform search (typically reinforcement learning but
also increasingly random search~\cite{DBLP:conf/uai/LiT19}). This has
led to exciting architectures like
EfficientNet~\cite{DBLP:conf/icml/TanL19}.\footnote{It is worth noting
  that data augmentation plays an important role in this
  architecture.} This is a tremendously exciting area with regular
workshops at all major machine learning conferences. Overton is
inspired by this area. On a technical level, the search used in
Overton is a coarser-grained search than what is typically done in
NAS. In particular, Overton searches over relatively limited large
blocks, e.g., should we use an LSTM or CNN, not at a fine-grained
level of connections. In preliminary experiments, NAS methods seemed
to have diminishing returns and be quite expensive. More sophisticated
search could only improve Overton, and we are excited to continue to
apply advances in this area to Overton. Speed of developer iteration
and the ability to ship production models seems was a higher priority
than exploring fine details of architecture in Overton.

\paragraph*{Statistical Relational Learning} 
Overton's use of a relational schema to abstract statistical reasoning
is inspired by Statistical Relational Learning (SRL), such as Markov
Logic~\cite{DBLP:journals/cacm/DomingosL19}. DeepDive~\cite{DBLP:journals/pvldb/ShinWWSZR15},
which is based on Markov Logic, allows one to wrap deep learning as
relational predicates, which could then be composed. This inspired
Overton's design of compositional payloads. In the terminology of
SRL~\cite{DBLP:journals/ftdb/BroeckS17}, Overton takes a knowledge
compilation approach (Overton does not have a distinct querying
phase). Supporting more complex, application-level constraints seems
ideally suited to an SRL approach, and is future work for Overton.

\section{Conclusion and Future Work}

This paper presented Overton, a system to help engineers manage the
lifecycle of production machine learning systems. A key idea is to use
a schema to separate the model from the supervision data, which allows
developers to focus on supervision as their primary interaction
method. A major direction of on-going work are the systems that build
on Overton to aid in managing data augmentation, programmatic
supervision, and collaboration.\\

{\noindent\textit{Acknowledgments} This work was made possible by
  Pablo Mendes, Seb Dery, and many others. We thank many teams in Siri
  Search, Knowledge, and Platform and Turi for support and
  feedback. We thank Mike Cafarella, Arun Kumar, Monica Lam, Megan
  Leszczynski, Avner May, Alex Ratner, Paroma Varma, Ming-Chuan Wu,
  Sen Wu, and Steve Young for feedback.}

{ \bibliographystyle{abbrv} \bibliography{overton, metal}}
\end{document}